# Linearization of ReLU Activation Function for Neural Network-Embedded Optimization: Optimal Day-Ahead Energy Scheduling

Cunzhi Zhao, Fan Jiang, and Xingpeng Li

*Abstract*— Recently, neural networks have been widely applied in the power system area. They can be used for better predicting input information and modeling system performance with increased accuracy. In some applications such as battery degradation neural network-based microgrid day-ahead energy scheduling, the input features of the trained learning model are variables to be solved in optimization models that enforce limits on the output of the same learning model. This will create a neural network-embedded optimization problem; the use of nonlinear activation functions in the neural network will make such problems extremely hard to solve if not unsolvable. To address this emerging challenge, this paper investigated different methods for linearizing the nonlinear activation functions with a particular focus on the widely used rectified linear unit (ReLU) function. Four linearization methods tailored for the ReLU activation function are developed, analyzed and compared in this paper. Each method employs a set of linear constraints to replace the ReLU function, effectively linearizing the optimization problem, which can overcome the computational challenges associated with the nonlinearity of the neural network model. These proposed linearization methods provide valuable tools for effectively solving optimization problems that integrate neural network models with ReLU activation functions.

*Index Terms*-- Day-ahead energy scheduling, Learning-embedded optimization, Linearization, Mixed-integer linear programming, Neural network, Rectified linear unit.

NOMENCLATURE

*Indices:*
| | |
|---|---|
| $g$ | Generator index. |
| $s$ | Battery energy storage system index. |
| $k$ | Transmission line index. |
| $l$ | Load index. |
| $wt$ | Wind turbine index. |
| $pv$ | Photovoltaic index. |

*Sets:*
| | |
|---|---|
| $T$ | Set of time intervals. |
| $G$ | Set of controllable micro generators. |
| $S$ | Set of energy storage systems. |
| $WT$ | Set of wind turbines. |
| $PV$ | Set of PV systems. |

*Parameters:*
| | |
|---|---|
| $c_g$ | Linear cost for controllable unit $g$. |
| $c_g^{NL}$ | No load cost for controllable unit $g$. |
| $c_g^{SU}$ | Start-up cost for controllable unit $g$. |
| $\Delta T$ | Length of a single dispatch interval. |
| $R_{prcnt}$ | Ratio of the backup power to the total power. |
| $E_s^{Max}$ | Maximum energy capacity of ESS $s$. |
| $E_s^{min}$ | Minimum energy capacity of ESS $s$. |
| $c_t^{Buy}$ | Wholesale electricity purchase price in time interval $t$. |
| $c_t^{Sell}$ | Wholesale electricity sell price in time interval $t$. |
| $P_g^{Max}$ | Maximum capacity of generator $g$. |
| $P_g^{Min}$ | Minimum capacity of generator $g$. |
| $P_k^{Max}$ | Maximum thermal limit of transmission line $k$. |
| $b_k$ | Susceptance, inverse of impedance, of branch $k$. |
| $P_{Grid}^{Max}$ | Maximum thermal limit of tie-line between main grid and microgrid. |
| $P_g^{Ramp}$ | Ramping limit of diesel generator $g$. |
| $P_s^{Max}$ | Maximum charge/discharge power of BESS $s$. |
| $P_s^{Min}$ | Minimum charge/discharge power of BESS $s$. |
| $\eta_s^{Disc}$ | Discharge efficiency of BESS $s$. |
| $\eta_s^{Char}$ | Charge efficiency of BESS $s$. |

*Variables:*
| | |
|---|---|
| $U_t^{Buy}$ | Status of buying power from main grid in time interval $t$. |
| $U_t^{Sell}$ | Status of selling power to main grid status in time $t$. |
| $U_{s,t}^{Char}$ | Charging status of energy storage system $s$ in time interval $t$. It is 1 if charging status; otherwise 0. |
| $U_{s,t}^{Disc}$ | Discharging status of energy storage system $i$ in time interval $t$. It is 1 if discharging status; otherwise 0. |
| $U_{g,t}$ | Status of generator $g$ in time interval $t$. It is 1 if on status; otherwise 0. |
| $V_{g,t}$ | Startup indicator of Status of generator $g$ in time interval $t$. It is 1 if unit $g$ starts up; otherwise 0. |
| $P_{g,t}$ | Output of generator $g$ in time interval $t$. |
| $\theta_{n(k)}^t$ | Phase angle of sending bus $n$ of branch $k$. |
| $\theta_{m(k)}^t$ | Phase angle of receiving bus $m$ of branch $k$. |
| $P_{k,t}$ | Line flow at transmission line $k$ at time period $t$. |
| $P_t^{Buy}$ | Amount of power purchased from main grid power in time interval $t$. |
| $P_t^{Sell}$ | Amount of power sold to main grid power in time interval $t$. |
| $P_{l,t}$ | Demand of the microgrid in time interval $t$. |
| $P_{s,t}^{Disc}$ | Discharging power of energy storage system $s$ at time $t$. |
| $P_{s,t}^{Char}$ | Charging power of energy storage system $s$ at time $t$. |

## I. INTRODUCTION

With the trend of decarbonization, a large number of renewable energy sources (RES)-based power plants are being constructed in the power system and the RES portfolio keeps increasing [1]. However, the intermittent and stochastic characteristics of the RES has raised the concern of system reliability and stability for high RES penetrated grid [2]. Deep learning (DL) is playing a crucial role in enhancing the operational efficiency and reliability of power systems, particularly in the context of high RES penetration and the associated challenges of intermittency and unpredictability [3]. DL has been widely adapted to assist the power system operations and reliability such as grid restoration [4], economic dispatch [5], and intermittent solar power forecasting [6]. DL is an important technology for contributing to the transition towards more sustainable and resilient power grids.

The outstanding performance of deep learning technologies has ushered in innovative solutions for numerous challenging power system issues that traditional methods struggle to address. Today's power system is confronting formidable challenges, primarily due to the extensive integration of RES into the grid. Deep learning methods have emerged as indispensable tools within the power system area. Based on different attributes of DL models, they have been found to be useful in diverse applications in solving a range of power system problems. For example, a deep neural network (DNN) consisting of fully-connected dense layers are developed in [7] to predict active power flows and it outperforms the widely-used linearized DC power flow model. The utilization of graph neural networks (GNN) in [8] allows for efficient predictions of current and power injection, capitalizing on GNN's topological advantages. Convolutional neural network (CNN) is adapted to predict the rate of change of frequency under large disturbances to ensure the system stability in [9]. Furthermore, fault detection including fault type and location can be predicted by the artificial neural network as presented in [10]. Meanwhile, recurrent neural networks are widely used in time sequential prediction such as electricity price prediction [11], load forecasting [12], weather forecasting [13], and renewable generation forecasting [14], offering versatile solutions to contemporary power system challenges.

DL is also widely applied in various optimization models for day-ahead generation scheduling and real-time dispatching, including optimal power flow (OPF) and unit commitment [15]-[19], to lighten computational burden and reduce solving time. A comprehensive review in [15] summarized recent ML applications for grid control. The literatures can be roughly divided into two categories: (1) DL is employed to directly solve optimal models; such as in [16] and [17], DNN are used to solve OPF, bypassing the traditional iterative procedures, making it more efficient for large-scale power grids; (2) DL is used as a part of optimal models to obtain constraints, for example, the authors in [18] utilize a back-propagation neural network to determine the security boundary, which serves as a constraint in security-constrained OPF; and in [19], a DNN-based frequency stability performance metric predictor is proposed and trained for estimating grid frequency response following largest generator outage, which is reformulated as linear equations with binary variables and then used as additional constraints in day-ahead unit commitment model to ensure grid frequency performance.

Most DL models adopt the rectified linear unit (ReLU) as the nonlinear activation function between the hidden layers to enhance the training efficiency and robustness [21]. Nevertheless, the ReLU introduces the nonlinearity to the DL models which make them nonlinear [22]. While this nonlinearity poses no issue for direct inference use like fault detection and power flow prediction, it can become a significant obstacle when integrating the trained DNN model into optimization problems where DNN's input features are decision variables to be solved, rendering them suddenly unsolvable due to the introduced nonlinearity. Remarkably, none of the previously mentioned studies have addressed the crucial challenge of linearizing the ReLU function in DNNs. Thus, there exists a significant research gap concerning the development of methods to linearize ReLU-based DNNs and thus accelerate the solving process for emerging DL-embedded optimization problems.

To address the challenge, we have proposed four different models to linearize the ReLU activation function. The proposed four models include: (i) Big-M based piecewise linearization (BPWL), (ii) convex triangle area relaxation (CTAR), (iii) penalized CTAR (P-CTAR), and (iv) penalized convex area relaxation (PCAR). BPWL is able to fully linearize the ReLU activation function without any approximation while the rest three models make some approximations to decrease the computational complexity.

Our previous work [23]-[24] has introduced a novel neural network based battery degradation (NNBD) model aimed at accurately quantifying the battery degradation values per usage profile. The proposed NNBD model is able to predict the battery degradation value for each cycle based on the input of state of charge (SOC), state of health (SOH), depth of discharge (DOD), ambient temperature, and charge/discharge rate (C Rate) [25]. The NNBD model enables the incorporation of battery degradation into microgrid daily operational energy scheduling. However, this integration encountered unexpected computational burden due to the nonlinear nature of the NNBD model that utilizes the ReLU activation function in the hidden layers; this poses challenges when NNBD is incorporated into the optimal day-ahead generation scheduling problems. To address this issue, we will evaluate the proposed four ReLU linearization models in the testbed of NNBD-integrated microgrid day-ahead scheduling (MDS) model. It is worth mentioning that the proposed ReLU linearization methods not only fit the proposed NNBD model and optimal energy scheduling applications, but also applicable to a broader spectrum of DL-embedded optimization models that contains neural networks with ReLU activation functions.

The main contributions of this paper are as follows:
- *Linearization Approaches*: we introduce four novel formulations for linearizing the ReLU activation function, enhancing its applicability in various contexts, particularly when the input features of DL models are variables to be solved by an optimization model while the outputs represent physical system performance subject to some requirements enforced in the same optimization model.

- *Linearized Energy Scheduling Frameworks*: building upon the linearized ReLU models, we develop four distinct day-ahead scheduling models, providing practical solutions for generation scheduling.
- *Performance Assessment*: Comprehensive evaluations of the linearized ReLU models demonstrate their effectiveness in addressing nonlinearity challenges.
- *Optimal Configuration Exploration*: we conduct sensitivity tests to identify optimal configurations for the proposed linearization models, ensuring their robust performance in diverse scenarios.

The rest of the paper is organized as follows. Section II describes the proposed ReLU linearization models. Section III presents the traditional day-ahead scheduling model. Section IV presents the neural network integrated day-ahead scheduling model. Section V presents case studies and Section VI concludes the paper.

## II. PROPOSED LINEARIZATION MODELS

This section presents the formulations of four models designed to linearize the ReLU activation function within the neural network model.

The fully connected neural network models are characterized by a series of equations that describe the calculation and activation processes of neurons. The pre-activated value $x_h^i$ of each neuron is computed by (1), factoring in input features $a_{h-1}^i$ from the previous layer, the corresponding weight matrix $W$, and biases matrix $B$. Most neural network models employ ReLU as the activation function, as shown in (2).

$$x_h^i = \sum a_{h-1}^i * W + B \quad (1)$$

$$a_h^i = ReLU(x_h^i) = max(0, x_h^i) \quad (2)$$

While this activation function is prevalent for introducing nonlinearity to capture intricate relationships among variables, the nonlinearity of the ReLU function can pose challenges when it is embedded in the optimization problem. To address this challenge, the ReLU activation function can be linearized by applying the proposed linearization models described in this section. Notably, the proposed linearization models can be applied to any optimizations models that needs to efficiently integrate nonlinear ReLU activation function.

### A. The BPWL method

BPWL method is adapted to reformulate the ReLU function represented by (1) into (3)-(6), and the illustration is shown in Fig.1, where $\delta_h^i$ is a binary variable represents the activation status of neuron $i$ in layer $h$, and $M$ is a pre-specified numerical value that is larger than any possible value of $|x|$.

$$a_h^i \leq x_h^i + M * (1 - \delta_h^i) \quad (3)$$
$$a_h^i \geq x_h^i \quad (4)$$
$$a_h^i \leq M * \delta_h^i \quad (5)$$
$$a_h^i \geq 0 \quad (6)$$

As shown in Fig. 1, the BPWL model offers the distinct advantage of perfectly linearizing the ReLU activation function without any reformulation losses, which is illustrated by the fact that the BPWL line in Fig. 1 completely overlaps the ReLU function curve. However, this method requires one additional binary variable for each neuron that applies the ReLU function, which may significantly increase the computational complexity.

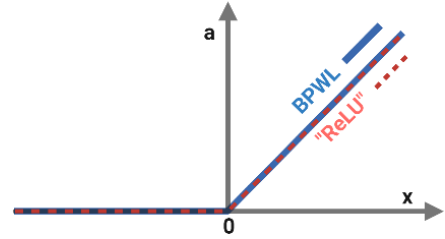

Fig. 1. Illustration of the BPWL model for ReLU linearization.

### B. The CTAR method

The proposed CTAR approximate the ReLU function at each neuron with (4) and (6)-(7), which constrains the feasible solution set, and the illustration is shown in Fig. 2. CTAR offers a pure linear representation of ReLU, introducing minimal complexity to optimization problems while acknowledging the presence of approximation errors.

$$a_h^i \leq \frac{UB}{UB - LB} x_h^i - \frac{UB \cdot LB}{UB - LB} \quad (7)$$

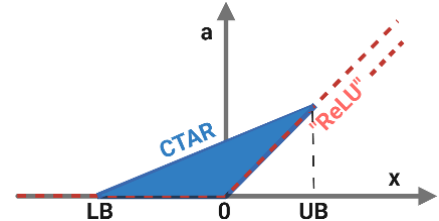

Fig. 2. Illustration of the CTAR model for ReLU linearization.

As shown in Fig. 2, the blue area denotes the feasible region, delimited by the lower bound (LB) and upper bound (UB), both of which are determinable from the neural network model. Specifically, LB should be less than the minimum neuron preactivated value, while UB should exceed the maximum neuron preactivated value. In most cases with normalized training data, each neuron's value falls within the range between -1 to 1. It's noteworthy that different choices for LB and UB can impact the performance of the CTAR linearization method.

### C. The P-CTAR method

The proposed P-CTAR model is introduced to reduce the approximation error associated with CTAR model with (4) and (6)-(8). To achieve this, a mitigation strategy involves incorporating a penalty term $c_h$ into the objective function of the optimization model. This penalty encourages the nonnegative variable $a_h^i$ to be positioned closer to the lower two sides, corresponding to the actual ReLU activated values, within the triangular representation as shown in Fig. 3. To minimize the approximation error, $a_h^i$ should be the minimum value while fulfilling (4), (6), and (7). Therefore, the set of $a_h^i$ that satisfies the requirement approximately falls on the line where the red dotted line meets the blue area.

$$f^c = \sum a_h^i c_h \qquad (8)$$

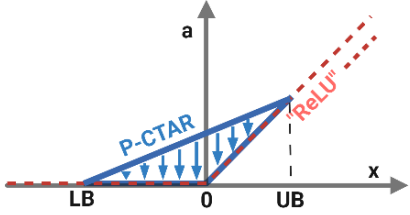

Fig. 3. Illustration of the P-CTAR model for ReLU linearization.

### D. The PCAR method

The proposed PCAR method is designed to add a penalty term $c_h$ on nonnegative $a_h^i$ without LB and UB, represented by (4), (6), and (8), which force the $a_h^i$ to be set at the ReLU activated values as shown in Fig. 4.

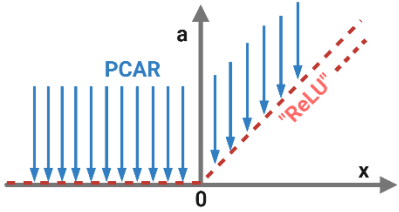

Fig. 4. Illustration of the PCAR model for ReLU linearization.

As compared to CTAR, the PCAR method offers enhanced accuracy, especially when equipped with sufficiently large penalty terms, and notable efficiency gains due to the absence of constraints (7). Thus, this method stands out for its ability to significantly reduce computational complexity when compared to other available linearization approaches.

## III. TRADITIONAL MICROGRID DAY-AHEAD ENERGY SCHEDULING MODEL

This section presents the traditional microgrid day-ahead scheduling model. This model consists of (9)-(26) as described below and it does not consider battery degradation.

The objective of this traditional MDS model is to minimize the total cost of the microgrid operations as illustrated in (9). The MDS model includes a power balance equation, as detailed in (10), encompassing controllable generators, RES, power exchange with the main grid, battery energy storage system (BESS) output, and the load. Constraint (11) enforces the power limits of the controllable units such as diesel generators, while (12) and (13) enforce the ramping up and down limits. Equations (14)-(16) are employed to establish the relationship between a controllable unit's start-up status and its on/off status. Equation (17) restricts the BESS to be either in charging mode or in discharging mode or stay idle. Constraints (18)-(19) limit the charging/discharging power of BESS. Equation (20) governs the power exchange status between the microgrid and the main grid, indicating whether it involves purchasing, selling, or remaining idle. Constraints (21)-(22) define the thermal limits of the tie-line. Equation (23) computes the energy stored in the BESS for each time interval. Constraint (24) ensures that the final energy of the BESS equals the initial energy value while (25) respects the BESS capacity limit. Constraint (26) guarantees the microgrid maintains sufficient backup power to handle outage events.

Objective function:

$$f^{MG} = \sum\sum (P_{g,t} c_g + U_{g,t} c_g^{NL} + V_{g,t} c_g^{SU}) \\ + P_t^{Buy} c_t^{Buy} - P_t^{Sell} c_t^{Sell}, \forall g, t \qquad (9)$$

Constraints are as follows:

$$P_t^{Buy} + \sum_{g \in S_g} P_{g,t} + \sum_{wt \in S_{wt}} P_{wt,t} + \sum_{pv \in S_{pv}} P_{pv,t} \\ + \sum_{s \in S_s} P_{s,t}^{Disc} = P_t^{Sell} + \sum_{l \in S_l} P_{l,t} + \sum_{s \in S_s} P_{s,t}^{Char} \qquad (10)$$

$$P_g^{Min} \le P_{g,t} \le P_g^{Max}, \forall g, t \qquad (11)$$

$$P_{g,t+1} - P_{g,t} \le \Delta T \cdot P_g^{Ramp}, \forall g, t \qquad (12)$$

$$P_{g,t} - P_{g,t+1} \le \Delta T \cdot P_g^{Ramp}, \forall g, t \qquad (13)$$

$$V_{g,t} \ge U_{g,t} - U_{g,t-1}, \forall g, t, \qquad (14)$$

$$V_{g,t+1} \le 1 - U_{g,t}, \forall g, t, \qquad (15)$$

$$V_{g,t} \le U_{g,t}, \forall g, t, \qquad (16)$$

$$U_{s,t}^{Disc} + U_{s,t}^{Char} \le 1, \forall s, t \qquad (17)$$

$$U_{s,t}^{Char} \cdot P_s^{Min} \le P_{s,t}^{Char} \le U_{s,t}^{Char} \cdot P_s^{Max}, \forall s, t \qquad (18)$$

$$U_{s,t}^{Disc} \cdot P_s^{Min} \le P_{s,t}^{Disc} \le U_{s,t}^{Disc} \cdot P_s^{Max}, \forall s, t \qquad (19)$$

$$U_t^{Buy} + U_t^{Sell} \le 1, \forall t \qquad (20)$$

$$0 \le P_t^{Buy} \le U_t^{Buy} \cdot P_{Grid}^{Max}, \forall t \qquad (21)$$

$$0 \le P_t^{Sell} \le U_t^{Sell} \cdot P_{Grid}^{Max}, \forall t \qquad (22)$$

$$E_{s,t} - E_{s,t-1} + \Delta T(P_{s,t-1}^{Disc}/\eta_s^{Disc} - P_{s,t-1}^{Char} \eta_s^{Char}) \\ = 0, \forall s, t \qquad (23)$$

$$E_{s,t=24} = E_s^{Initial}, \forall s \qquad (24)$$

$$0 \le E_{s,t} \le E_{s,t}^{max} \qquad (25)$$

$$P_{Grid}^{Max} - P_t^{Buy} + P_t^{Sell} + \sum_{g \in S_g}(P_g^{Max} - P_{g,t}) \\ \ge R_{percent}\left(\sum_{l \in S_l} P_{l,t}\right), \forall t \qquad (26)$$

## IV. NEURAL NETWORK-INTEGRATED OPTIMAL DAY-AHEAD SCHEDULING MODEL

This section presents a neural network-integrated optimal day-ahead scheduling model which is linearized by the four linearization methods proposed in Section II.

### A. Neural Network Based Battery Degradation Model

We have constructed a fully connected neural network model as mentioned in Section I to predict battery degradation, with five key aging factors (ambient temperature, C Rate, SOC, DOD, and SOH) comprising a five-element input vector for the network. Each input vector corresponds to a single output value, representing the battery degradation as a percentage relative to the SOH for the corresponding cycle.

The NN model used in this paper is an open-source model [26]. The dataset is obtained from a battery aging test model developed in MATLAB Simulink [27]. This work conducted 945 different battery aging tests with different values of degradation factors. The data are pre-processed by the normalization method. The dataset is split as 80% training dataset and 20% validation dataset. The loss function during the training process is mean square error (MSE).

The structure of the trained neural network is shown in Fig. 5 [22] plotted by NN-SVG. The NNBD model has an input layer with 5 neurons corresponding to the 5 input features, first hidden layer with 20 neurons, second hidden layer with 10 neurons and an output layer with 1 neuron indicating the percentage battery degradation. The activation function is ReLU for the hidden layers and "linear" for the output layer.

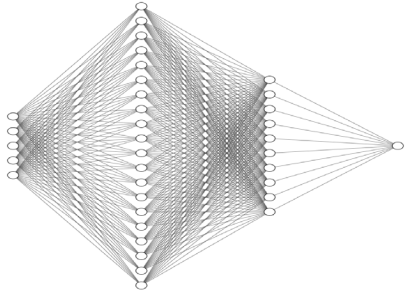

Fig. 5. Structure of the NNBD model [22].

A comparison between one-layer and two-layer NNBD models is done to determine the number of layers of the NNBD model. The training and validation loss of the two NN models are compared in Table I. This one-layer NNBD model has 5 inputs, 1 output, which is the same as the two-layer NNBD model. The hidden layer has 20 neurons, and the activation function is ReLU. The results show that the two-layer model architecture has the better performance; therefore, the two-layer NN model is selected and used in this paper.

Table I Results of One-Layer and Two- Layer NNBD models

| NNBD Models | One-layer | Two-layer |
|---|---|---|
| Average loss | 0.858% | 0.816% |
| Average error between actual values and predicted value | 6.78% | 4.88% |
| Accuracy of 10% error tolerance | 22.98% | 18.57% |
| Accuracy of 15% error tolerance | 9.66% | 3.47% |
| Accuracy of 20% error tolerance | 2.07% | 1.08% |

### B. NNBD based Day-ahead Scheduling Model

Based on the BESS operation profile, the SOC level is required as the input of the proposed NNBD model. DOD is calculated by taking the absolute difference in SOC levels between time intervals $t$ and $t$-1, as shown in (27). C rate is calculated by (28). The input vector can be formed as shown in (29) and then fed into the trained NNBD model to obtain the total battery degradation over the MDS time horizon in (30). The sum of the battery degradation is used to calculate the equivalent battery degradation cost by (31). The updated objective function (32) is required for two of the proposed models: BPWL and CTAR. The NNBD based day-ahead scheduling model will consists of (9)-(32).

$$DOD_t = |SOC_t - SOC_{t-1}| \quad (27)$$
$$C_t^{Rate} = DOD_t/\Delta T \quad (28)$$
$$\overline{x}_t = (T, C, SOC, DOD, SOH) \quad (29)$$
$$BD = \sum_{t \in S_T} f^{NN}(\overline{x}_t) SOH \quad (30)$$
$$f^{BESS} = \frac{c_{BESS}^{Capital} - c_{BESS}^{SV}}{1 - SOH_{EOL}} BD \quad (31)$$
$$f = f^{MG} + f^{BESS} \quad (32)$$

However, due to the nonlinearity of the NNBD model, especially due to the ReLU activation function, the integration of the proposed NNBD model into the day-ahead scheduling framework cannot be directly solved. Therefore, we employ the proposed linearization methods within the intricate optimization model to facilitate its efficient solution.

### C. Linearization for NNBD-based MDS Model

Since the nonlinear ReLU activation function make the optimization model extremely hard to solve if not unsolvable, we have reformulated the MDS models to make it linear and solvable with the proposed four ReLU linearization models respectively. The updated objective function (33) is required for two of the proposed models: P-CTAR and PCAR. The linearized NNBD-integrated day-ahead energy scheduling models are defined in Table II.

$$f = f^{MG} + f^{BESS} + f^c \quad (33)$$

Table II The Proposed Linearized Day-Ahead Scheduling Models

| Models | Equations |
|---|---|
| BPWL-MDS | (3)-(6), (9)-(32) |
| CTAR-MDS | (4), (6)-(7), (9)-(32) |
| P-CTAR-MDS | (4), (6)-(8), (9)-(31), (33) |
| PCAR-MDS | (4), (6), (8), (9)-(31), (33) |

## V. CASE STUDIES

### A. Microgrid Testbed

To evaluate the effectiveness of the linearized day-ahead scheduling model, we utilized a representative grid-connected microgrid [28] as the test case in this work, featuring several distributed energy resources including renewables and controllable units. The microgrid setup encompasses various key components, notably a conventional diesel generator, wind turbines, residential houses with solar panels, and a lithium-ion BESS with a charging and discharging roundtrip efficiency of 90%. The parameters for these main components are provided in Table III.

TABLE III MICROGRID TESTBED

| Main Components | Diesel Generator | Wind Turbines | Solar Panels | Lithium-ion BESS |
|---|---|---|---|---|
| Size | 180kW | 1000kW | 1500kW | 300kW |

The load and renewable energy profile are shown in Fig. 6. To simulate real-world scenarios accurately, the load data for the microgrid is obtained from the electricity consumption patterns of 1000 residential households from the Pecan Street Dataport [29]. Additionally, for a comprehensive representation of environmental conditions, the ambient temperature and available solar power data are sourced over a 24-hour period from the same Pecan Street Dataport source.

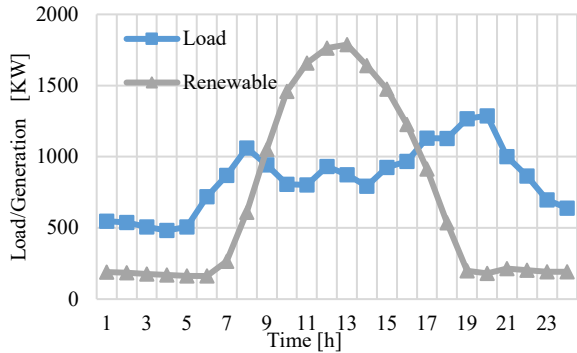

Fig. 6. Microgrid load profile.

Furthermore, the wholesale electricity price data used in the microgrid model is obtained from the Electric Reliability Council of Texas (ERCOT) that manages majority of the Texas grid [28].

### B. Results of Testbed

The MDS optimization problem in this paper was solved on a computer equipped with the following hardware: an AMD Ryzen 7 3800X processor, 32 GB of RAM, and an Nvidia Quadro RTX 2700 Super GPU with 8 GB of memory. The Pyomo [30] package, a powerful power system optimization modeling framework, was utilized to formulate and solve the day-ahead optimization problem. To expedite our quest for optimal solutions, we employed the high-performance mathematical programming solver Gurobi [31].

The value of M multiplier in BPWL method should be set to a suitable value. Both too large or too small M value may increase the solving time. To obtain the suitable value of M, Table IV shows the MDS results with different M values. It shows that we can obtain the least solving time when the M value is 100.

Table IV MDS Results with different M values

| M values | 10000 | 1000 | 100 | 10 |
|---|---|---|---|---|
| Degradation | 0.0365% | 0.0365% | 0.0365% | 0.0365% |
| Total Cost | $529.50 | $529.50 | $529.50 | $529.50 |
| Solving Time | 33.03s | 14.64s | 12.27s | 26.94s |

Table V presents the performance results obtained from the microgrid testbed based on the four proposed linearization models. In Table V, the term, *Degradation*, represents the degradation value calculated by the proposed linearization methods, while the term, *Real Degradation*, represents the degradation value computed separately by the NNBD model with the BESS operation profile to gauge the accuracy of the linearization models. The term, *Error*, is derived from the comparison between *Degradation* and *Real Degradation*. The term, *Total Cost*, represents the combined MDS operation cost and the equivalent degradation cost. It's important to note that this equivalent degradation cost is derived from *Degradation* which may introduce the linearization approximation error. The term, *Real Cost* represent the combined MDS operation cost and the real equivalent degradation cost based on the *Real Degradation*. Also, the *Total Cost* in Table V has already excluded the penalty cost for P-CTAR and PCAR models.

Table V Results of NNBD-MDS with various linearization methods

| MDS | BPWL | CTAR | P-CTAR | PCAR |
|---|---|---|---|---|
| Degradation | 0.0365% | 0.0083% | 0.0450% | 0.0945% |
| Real Degradation | 0.0375% | 0.0471% | 0.0408% | 0.1546% |
| Error % | 2.62% | 82.28% | 10.18% | 38.87% |
| Total Cost | $529.50 | $488.84 | $552.09 | $651.87 |
| Real Cost | $530.68 | $535.32 | $547.11 | $723.99 |
| Solving Time | 12.27 s | 0.27 s | 2.76 s | 0.49 s |

The outcomes indicate that BPWL yields the lowest linearization error, attributed to its complete linearization of the ReLU activation function. The minimal degradation prediction error is a result of rounding the weights in the NNBD model during the battery degradation value verification process. Ideally, there should be zero linearization error.

However, it's worth mentioning that the BPWL model does require a relatively longer solving time due to the exponential addition of new binary variables to the optimization problem. Although the current solving time of 12.27 seconds remains acceptable for a microgrid case, it will increase significantly for larger systems. Overall, the BPWL model accurately represents the ReLU function with no reformulation losses but introduces binary variables, complicating the optimization model and leading to longer solving times. In contrast, the computation times for the other linearization methods are significantly less. Notably, PCAR yields the shortest solving time, as expected, given its fewer constraints compared to CTAR and P-CTAR.

The three MDS models with non-exact ReLU linearization achieve different accuracies. The proposed P-CTAR model exhibits the second-lowest error among the four proposed methods, signifying its ability to accurately linearize the ReLU activation function. A comparative analysis between CTAR and P-CTAR reveals that CTAR introduces substantial linearization errors, suggesting an ineffective linearization approach in this particular case. The significant approximation error in CTAR results from the absence of a penalty term in the model. This may change if the MDS model changes. In contrast, the P-CTAR model incorporates a penalty term in addition to the CTAR model, leading to a considerable enhancement in performance relative to CTAR. The P-CTAR model outperforms the PCAR model which implies that the chosen lower bound and upper bound values for CTAR are appropriate and effective in this testbed.

The PCAR model, however, exhibits a linearization error of 38.87%, notably higher than P-CTAR. It's important to note that both PCAR and P-CTAR models' linearization errors are influenced by the penalty constant in the objective function, which requires additional sensitivity test to determine the optimal setup. While BPWL outperforms the other three models in terms of linearization accuracy, it lags behind in solving efficiency, especially when considering larger systems where it could potentially lead to significantly extended solving times.

In contrast, P-CTAR stands out as the best performance model in terms of linearization error among the rest of the models. The solving time is decreased significantly compared to the BPWL model while maintaining the linearization performance.

The linearization performance of the P-CTAR model when compared to the reference NNBD degradation value is illustrated graphically in Fig. 7, which is based on the BESS operation profile per usage cycle.

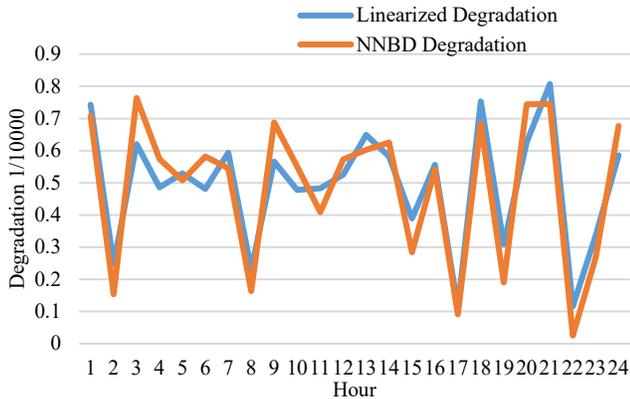
Fig. 7. P-CTAR model degradation comparison.

It can be seen in Fig. 7 that the linearized degradation, produced directly from the optimization model using the P-CTAR model to linearize the internal ReLU function of the NNBD model, exhibits a trend that closely aligns with the benchmark model. While there are some deviations in specific time periods, the overall degradation value over the 24-hour look-ahead scheduling horizon demonstrates minimal prediction error, affirming the efficacy of the P-CTAR linearization model.

It's noteworthy that during the design of the proposed linearization models, the expectation was that P-CTAR might introduce additional complexity to the optimization model, potentially leading to longer solving times. The results indeed align with this expectation, as the P-CTAR model's solving time surpasses that of the CTAR model. This trade-off between model efficiency and accuracy is duly considered in our evaluation.

## C. Sensitive Analysis of Penalty Cost in P-CTAR and CTAR

The results of a sensitivity test for the P-CTAR model using different penalty cost constants denoted as $c_h$ are presented in Fig. 8. This penalty cost serves the purpose of mitigating the approximation error associated with the CTAR model. However, it's important to note that different values of $c_h$ may lead to different outcomes, as the penalty cost is directly integrated into the objective function of the optimization problem. The objective cost reflects the expenses incurred by the objective function (33), while the penalty cost accounts for the costs introduced by the P-CTAR model. The real cost represents the MDS operating cost and the battery degradation cost, calculated as the difference between the objective cost and the penalty cost. However, the battery degradation cost here may include an error due to the linearization approximation of the P-CTAR model.

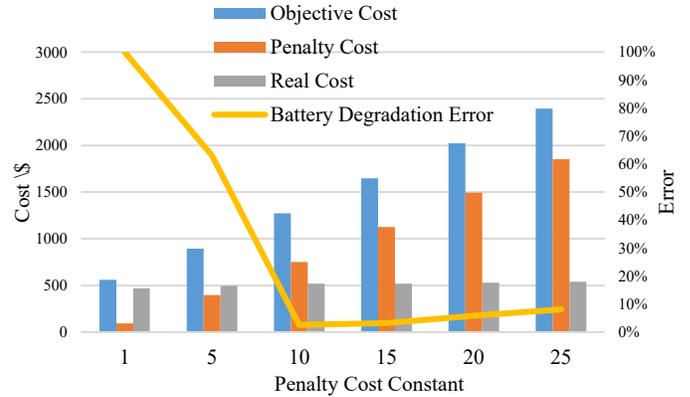
Fig. 8. P-CTAR sensitivity tests.

Upon examination of the figure, it becomes evident that there is no significant variation in the real cost as $c_h$ increases. However, the lowest linearization error is observed when $c_h$ is set to 10. Beyond this threshold, increasing $c_h$ leads to an escalation in battery degradation error which is directly caused by the error of linearization model. In summary, the selection of $c_h$ demands careful consideration and thorough preliminary testing to ensure the optimal performance of the P-CTAR model. It's essential to emphasize that the optimal $c_h$ value provided here pertains specifically to the NNBD-based MDS optimization problem. If the optimization model undergoes any modifications, it would require recalibration to determine the ideal $c_h$ value for the optimal setup.

We also conducted a sensitivity test on the PCAR model with varying $c_h$ values, and the results are depicted in Fig. 9.

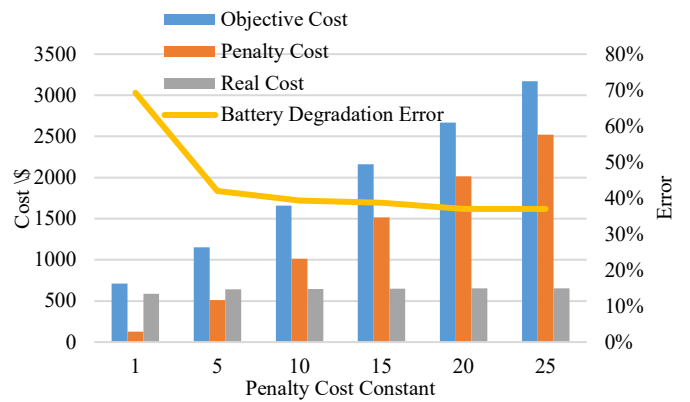
Fig. 9. PCAR sensitivity tests.

In Fig.9, it is noticeable that as we increase the $c_h$ value, the linearization error does not exhibit a substantial decrease. In essence, there doesn't appear to be an optimal $c_h$ value for the PCAR model that significantly enhances linearization accuracy. While the PCAR model demonstrates effectiveness, it doesn't achieve the same level of performance as the P-

CTAR model. Consequently, for the NNBD-integrated MDS optimization problem, the PCAR model may not represent the most ideal solution. However, it's worth noting that the PCAR model could potentially serve as the optimal solution for other neural network-based optimization models with distinct characteristics and requirements.

*D. Sensitive Analysis of RES Levels*

As the global decarbonization goal requires substantial increase of RES penetration, it is necessary to clarify how the P-CTAR and PCAR models will perform under different RES levels, and determine whether they are still good choices considering both solving time and the accuracy of the results.

Fig. 10 shows the results of total cost, real cost and battery degradation error of P-CTAR model with different RES levels, and Fig. 11 illustrates those of PCAR model.

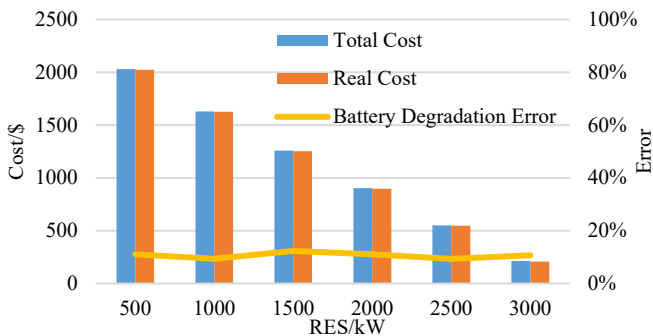

Fig. 10 P-CTAR sensitive test of RES levels.

It can be seen in Fig. 10 that the battery degradation errors with different RES levels are all around 10%. The total cost and real total cost decrease when RES level increases, because the microgrid needs to buy less power from power grids. Additionally, the total costs under different RES levels are almost the same as the real total cost, which are consistent with the results shown in Table V. Therefore, the P-CTAR model is still applicable when RES level changes.

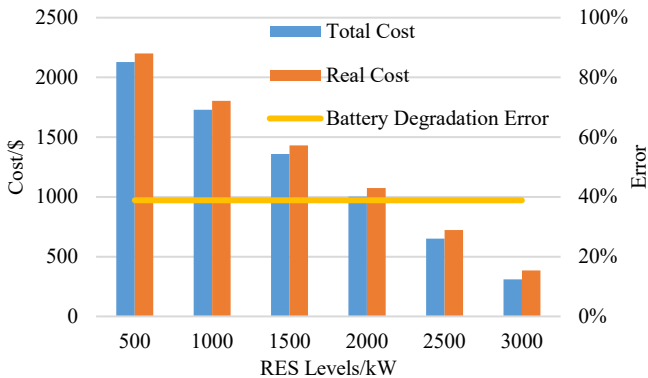

Fig. 11 PCAR sensitive test of RES levels.

The results in Fig. 11 have the same trend with those in Fig. 10, and battery degradation errors with different RES levels are all around 38%. Considering its fast-solving time, the PCAR model would still be useful in heavy-computational cases.

## VI. CONCLUSION

This paper investigated four innovative linearization models designed to tackle the challenges caused by the ReLU activation function in neural networks when integrated into optimization models where a subset of decision variables serve as the input features of the learning model. The inherent non-linearity of the ReLU function often makes such models complex and difficult to solve directly. However, by harnessing the proposed linearization models BPWL, CTAR, P-CTAR, and PCAR, it becomes feasible to effectively address the intricacies of neural network-integrated optimization problems. Our findings demonstrate that the BPWL model achieves the best performance in terms of accuracy by fully reformulating the ReLU activation function with auxiliary binary variables and it can achieve correct problem solutions. Furthermore, the proposed P-CTAR model can maintain impressive linearization accuracy without significant compromise, while significantly reducing the computing time.

Crucially, our results highlight that the choice of penalty terms is pivotal in obtaining optimal solutions for P-CTAR and PCAR models. While CTAR and PCAR may not deliver perfect performance in the NNBD-integrated microgrid testbed, they remain viable options for modeling and solving this class of problems. It's important to emphasize that when changes are made to the optimization model, the performance of the proposed linearization models should be re-evaluated.

In summary, the linearization models introduced in this paper open up fresh avenues for efficiently addressing the challenges posed by nonlinear neural network-integrated optimization problems. In future work, we intend to delve deeper into how these linearization models influence optimization models, and to further develop linearization models tailored to other activation functions such as the softmax function.

## VII. FUTURE WORK

While this paper studies the reformulation and linearization of commonly-used ReLU as activation function, a potential impactful future work would be to study other commonly-used activation functions such as hyperbolic tangent and softmax function. To better capture variations and handle uncertainties from the load and renewable power generation, a scenario-based stochastic optimization could be another potential future work. While the application tested in this paper is a microgrid system, this work could be extended to a bulk grid system when network constraints need to be included, which is a much more complicated problem, to better show the effectiveness and scalability of the proposed linearization methods.

REFERENCES

[1] T. Mai et al., "Renewable Electricity Futures for the United States," *IEEE Transactions on Sustainable Energy*, vol. 5, no. 2, pp. 372-378, April 2014.
[2] B. Li, T.Chen, X. Wang and G. Giannakis, "Real-Time Energy Management in Microgrids With Reduced Battery Capacity Requirements," IEEE Trans. Smart Grid., vol. 10, no. 2, pp. 1928–1938, Mar. 2019.
[3] W. Liao, B. Bak-Jensen, J. R. Pillai, Y. Wang and Y. Wang, "A Review of Graph Neural Networks and Their Applications in Power Systems,"

Journal of Modern Power Systems and Clean Energy, vol. 10, no. 2, pp. 345-360, March 2022.
[4] A. S. Bretas and A. G. Phadke, "Artificial neural networks in power system restoration," IEEE Transactions on Power Delivery, vol. 18, no. 4, pp. 1181-1186, Oct. 2003.
[5] T. Yalcinoz and M. J. Short, "Neural networks approach for solving economic dispatch problem with transmission capacity constraints," IEEE Transactions on Power Systems, vol. 13, no. 2, pp. 307-313, May 1998.
[6] A. A. Shah, K. Ahmed, X. Han and A. Saleem, "A Novel Prediction Error-Based Power Forecasting Scheme for Real PV System Using PVUSA Model: A Grey Box-Based Neural Network Approach," IEEE Access, vol. 9, pp. 87196-87206, 2021.
[7] T. Pham and X. Li, "Reduced Optimal Power Flow Using Graph Neural Network," 2022 North American Power Symposium (NAPS), Salt Lake City, UT, USA, 2022, pp. 1-6.
[8] J. B. Hansen, S. N. Anfinsen and F. M. Bianchi, "Power Flow Balancing With Decentralized Graph Neural Networks," IEEE Transactions on Power Systems, vol. 38, no. 3, pp. 2423-2433, May 2023.
[9] Mingjian Tuo and Xingpeng Li, "Convolutional Neural Network-based RoCoF-Constrained Unit Commitment", arXiv, Aug. 2023.
[10] J. Gracia, A. J. Mazon and I. Zamora, "Best ANN structures for fault location in single-and double-circuit transmission lines," IEEE Transactions on Power Delivery, vol. 20, no. 4, pp. 2389-2395, Oct. 2005.
[11] Vasudharini Sridharan, Mingjian Tuo and Xingpeng Li, "Wholesale Electricity Price Forecasting using Integrated Long-term Recurrent Convolutional Network Model", Energies, 15(20), 7606, Oct. 2022.
[12] D. Ageng, C. -Y. Huang and R. -G. Cheng, "A Short-Term Household Load Forecasting Framework Using LSTM and Data Preparation," IEEE Access, vol. 9, pp. 167911-167919, 2021.
[13] Y. Yu, J. Cao and J. Zhu, "An LSTM Short-Term Solar Irradiance Forecasting Under Complicated Weather Conditions," IEEE Access, vol. 7, pp. 145651-145666, 2019.
[14] V. Suresh, F. Aksan, P. Janik, T. Sikorski and B. S. Revathi, "Probabilistic LSTM-Autoencoder Based Hour-Ahead Solar Power Forecasting Model for Intra-Day Electricity Market Participation: A Polish Case Study," IEEE Access, vol. 10, pp. 110628-110638, 2022.
[15] L. Duchesne, E. Karangelos, and L. Wehenkel, "Recent Developments in Machine Learning for Energy Systems Reliability Management," Proceedings of IEEE, vol. 108, no. 9, pp. 1656–1676, Sep. 2020.
[16] X. Pan, T. Zhao, M. Chen, and S. Zhang, "DeepOPF: A Deep Neural Network Approach for Security-Constrained DC Optimal Power Flow", IEEE Transactions on Power Systems, vol. 36, no. 3, pp. 1725–1735, May 2021.
[17] A. Stratigakos, S. Pineda, J. M. Morales, et al., "Interpretable Machine Learning for DC Optimal Power Flow With Feasibility Guarantees", IEEE Transactions on Power Systems, vol. 39, no. 3, pp. 5126–5137, May 2024.
[18] V. J. Gutierrez-Martinez, C. A. Cañizares, C. R. Fuerte-Esquivel, A. Pizano-Martinez, and X. Gu, "Neural-Network Security-Boundary Constrained Optimal Power Flow", IEEE Transactions on Power Systems, vol. 26, no. 1, pp. 63–72, Feb. 2011.
[19] Mingjian Tuo and Xingpeng Li, "Security-Constrained Unit Commitment Considering Locational Frequency Stability in Low-Inertia Power Grids", IEEE Transactions on Power Systems, vol. 38, no. 5, pp. 4134-4147, Sep. 2023.
[20] S. Dittmer, E. J. King and P. Maass, "Singular Values for ReLU Layers," IEEE Transactions on Neural Networks and Learning Systedms, vol. 31, no. 9, pp. 3594-3605, Sept. 2020.
[21] G. Wang, G. B. Giannakis and J. Chen, "Learning ReLU Networks on Linearly Separable Data: Algorithm, Optimality, and Generalization," IEEE Transactions on Signal Processing, vol. 67, no. 9, pp. 2357-2370, 1 May1, 2019.
[22] C. Zhao and X. Li, "Microgrid Optimal Energy Scheduling Considering Neural Network based Battery Degradation", IEEE Transactions on Power Systems, vol. 39, no. 1, pp. 1594-1606, Jan. 2024.
[23] Cunzhi Zhao, Xingpeng Li, and Yan Yao, "Quality Analysis of Battery Degradation Models with Real Battery Aging Experiment Data", Texas Power and Energy Conference, College Station, TX, USA, Feb. 2023.
[24] C. Zhao and X. Li, "An Alternative Method for Solving Security-Constraint Unit Commitment with Neural Network Based Battery Degradation Model", 54th North American Power Symposium, Salt Lake City, UT, USA, Oct. 2022.
[25] Cunzhi Zhao and Xingpeng Li, "A Novel Real-Time Energy Management Strategy for Grid-Supporting Microgrid: Enabling Flexible Trading Power", IEEE PES General Meeting, (Virtually), Washington D.C., USA, Jul. 2021.
[26] Cunzhi Zhao, Xingpeng Li. Microgrid Optimal Energy Scheduling with Battery Degradation. figshare. Dataset, 2023. https://doi.org/10.6084/m9.figshare.21959582.v1
[27] N. Omar et al., "Lithium iron phosphate based battery—Assessment of the aging parameters and development of cycle life model," Appl. Energy, vol. 113, pp. 1575–1585, Jan. 2014.
[28] "Dataport Resources," May, 2019 [online] Available: https://dataport.pecanstreet.org/academic.
[29] "ERCOT, Electric Reliability Council of Texas," [Online]. Available: http://www.ercot.com/.
[30] "Pyomo, Python Software packages.," Available: [Online]. Available: http://www.pyomo.org/.
[31] "Gurobi Optimization, Linear Programming Solver," [Online]. Available: https://www.gurobi.com/